\documentclass[letterpaper, 10 pt, conference]{ieeeconf}  
\usepackage{adjustbox}
\usepackage{amsmath}
\usepackage{amssymb}
\usepackage{booktabs}
\usepackage[style=base]{caption}
\usepackage{comment}
\usepackage{gensymb}
\usepackage{graphicx}
\usepackage{mathtools}
\usepackage{multirow}
\usepackage{multicol}
\usepackage[percent]{overpic}
\usepackage{pifont}
\usepackage{tabularx}
\usepackage{xspace}
\usepackage{xfrac}
\usepackage{wrapfig,lipsum}
\usepackage[table]{xcolor}
\usepackage{subcaption}

\definecolor{lightgray}{RGB}{131, 126 114}
\definecolor{darkgreen}{RGB}{0, 150, 0}
\definecolor{darkred}{RGB}{200, 0, 0}
\definecolor{darkblue}{RGB}{0, 0, 200}
\definecolor{gray}{RGB}{142,142,142}

\usepackage{xcolor}
\usepackage{soul}
\soulregister\cite7
\soulregister\citeRest7
\soulregister\ref7
\soulregister\pageref7

\makeatletter
\let\NAT@parse\undefined
\DeclareRobustCommand\onedot{\futurelet\@let@token\@onedot}
\def\@onedot{\ifx\@let@token.\else.\null\fi\xspace}

\makeatother

\usepackage[colorlinks,pagebackref=false,citecolor=blue,bookmarks=false,hypertexnames=true]{hyperref}

\begin{document}

\title{BEVDetNet: Bird's Eye View LiDAR Point Cloud based \\ Real-time 3D Object Detection for Autonomous Driving}

\author{Sambit Mohapatra$^{1}$, Senthil Yogamani$^{2}$, Heinrich Gotzig$^{1}$, Stefan Milz$^{3}$, and Patrick M\"ader$^{4}$ \\
{\normalsize 
$^{1}$Valeo Germany\hspace{0.3cm} 
$^{2}$Valeo Ireland \hspace{0.3cm} 
$^{3}$SpleenLab.ai, Germany \hspace{0.3cm}
$^{4}$TU Ilmenau, Germany
} 
}

\maketitle



\begin{abstract}
3D object detection based on LiDAR point clouds is a crucial module in autonomous driving particularly for long range sensing. Most of the research is focused on achieving higher accuracy and these models are not optimized for deployment on embedded systems from the perspective of latency and power efficiency. For high speed driving scenarios, latency is a crucial parameter as it provides more time to react to dangerous situations.  Typically a voxel or point-cloud based 3D convolution approach is utilized for this module. Firstly, they are inefficient on embedded platforms as they are not suitable for efficient parallelization. Secondly, they have a variable runtime due to level of sparsity of the scene which is against the determinism needed in a safety system. In this work, we aim to develop a very low latency algorithm with fixed runtime. 
We propose a novel semantic segmentation architecture as a single unified model for object center detection using key points, box predictions and orientation prediction using binned classification in a simpler Bird's Eye View (BEV) 2D representation. The proposed architecture can be trivially extended to include semantic segmentation classes like road without any additional computation.
The proposed model has a latency of 4 ms on the embedded Nvidia Xavier platform. The model is 5X faster than other top accuracy models with a minimal accuracy degradation of 2\% in Average Precision at IoU=0.5 on KITTI dataset.

\end{abstract}

\section{Introduction}

Typically, object detection algorithms for autonomous driving are focused on improving the detection accuracy \cite{arnold2019survey}. In this pursuit of better performance in accuracy, other embedded deployment metrics such as latency, memory footprint and computational complexity have often been neglected. From a safety and real-time perspective, latency of detection is an important metric particularly for high speed driving scenarios. Difference of milliseconds can play a crucial role in dangerous situations. Low latency detection can lead to faster collision avoidance maneuvers and enables human drivers to react to these scenarios in Level 3 systems. Machine learning models are finally deployed on a low-power embedded Electronics Control Units (ECU) where memory footprint and computational complexity are also key aspects. Information from High definition maps can improve efficiency by subtracting out static infrastructure information \cite{ravi2018real}. Efficient fusion architectures \cite{rashed2019fusemodnet,el2019rgb} combining information from other sensors can also improve system efficiency.
These system level optimizations still need an efficient low latency model which is the focus of this paper.

Conventional methods for object detection from LiDAR 3D point clouds either use 3D convolutions or use techniques like voxelization for creating a grid-based representation of the irregular point clouds and then predict objects in these voxels. 3D convolutions are expensive and are not suitable for real-time performance. There are also other instances of converting point clouds into pseudo images of lesser dimensions using concepts like occupancy grids such as PointPillars \cite{lang2019pointpillars}. It improves computational efficiency significantly but they have introduced an additional complex pre-processing step. Another common practice in object detection models is the use of anchor boxes and non-maximum suppression steps. They are compute and memory intensive particularly in 3D space. {Complex-YOLO \cite{simony2018complex} is one of the first real-time models which uses Euler regression approach for orientation regressing using complex numbers. }

\begin{figure}[t]
    \centering
    \includegraphics[trim={1.5cm 1cm 1cm 0},clip,width=0.47\textwidth]{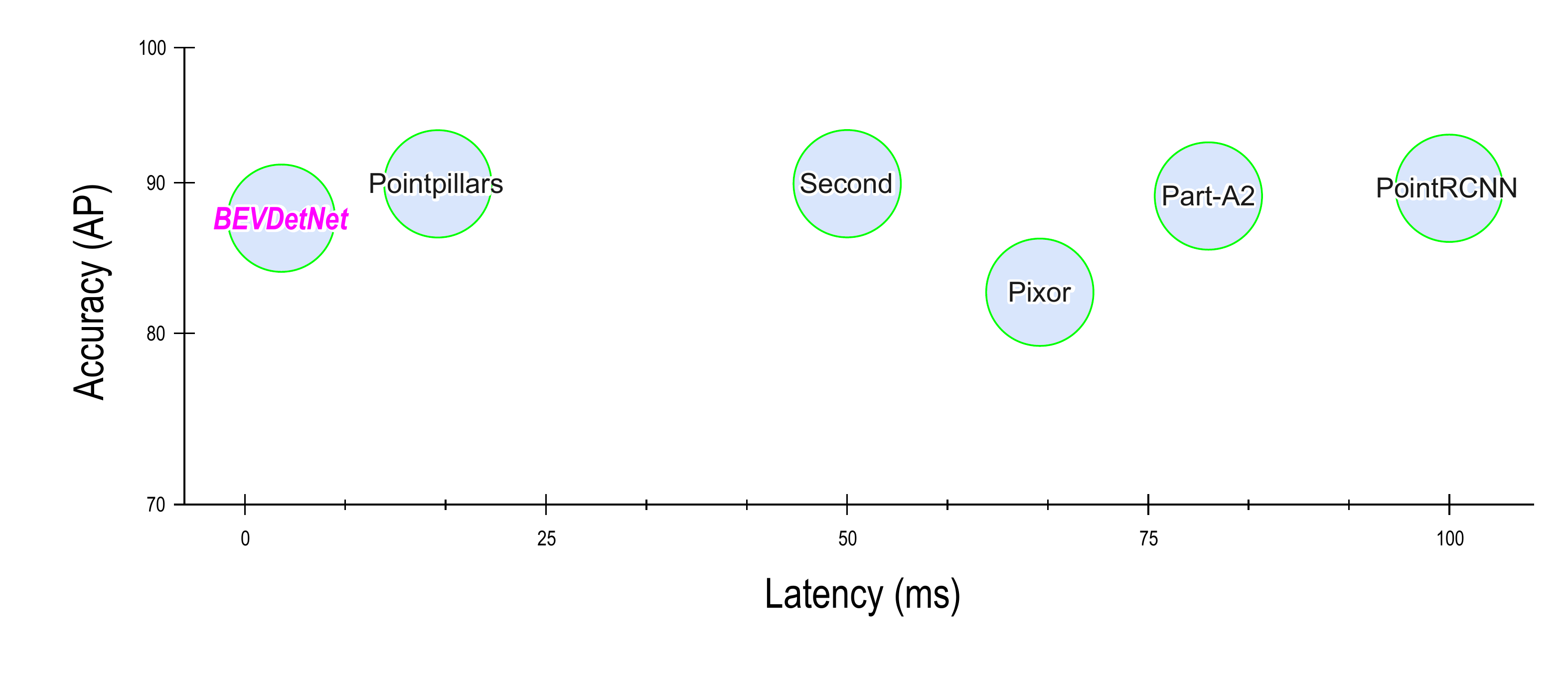}    
    \caption{Comparison of our BEVDetNet in terms of accuracy using IoU=0.5 and computation latency summarized from Table \ref{tab:Table_AP_latency}. Our method is 5X faster than the closest algorithm and it only has 2\% lesser accuracy than the state-of-the-art. 
    }
    \vspace{-5mm}
    \label{fig:Figure_AP_latency_compare}
\end{figure}

In this paper, we aim to design a low-latency model for 3D object detection deployable on an automotive embedded system. We prioritized the embedded constraints and then analyzed the accuracy impact. Our model \textit{BEVDetNet} achieves high performance with minimal degradation in accuracy as shown in Figure~\ref{fig:Figure_AP_latency_compare}. Our model is at least 5X faster than the closest real-time model with a minimal degradation of 2\% accuracy.
We make use of simpler 2D Bird’s  Eye View (BEV)  representation that allows us to employ fast and efficient 2D convolutions and avoids the computational burdens associated with 3D convolutions and other intermediate representations. We propose a simple segmentation architecture which outputs object centers as key points, box predictions {as regression output} and orientation classification as bins. We carefully design an efficient architecture using a hybrid ResNet and dilated convolutions architecture. We perform ablation studies to identify the optimal architecture.

\noindent Summary of our contributions include: 
\begin{itemize}
    \item A very low-latency model achieving 2 ms on Nvidia Xavier embedded platform.
    \item A unified segmentation architecture which jointly performs center regression and oriented box prediction.
    \item Extensive experimentation and ablation studies on KITTI dataset. 
\end{itemize}

\section{Related Work}

\subsection{3D object detection in LiDAR point clouds}
First successful attempts to object detection using point clouds, such as VoxelNet \cite{zhou2018voxelnet}, divide the point cloud into 3D volumes called voxels and then apply 3D convolutions for feature extraction. However, this is very computationally intensive and results in high latency. Frustum PointNets \cite{qi2018frustum}  uses a 2D detector for region proposals in camera images and then collects corresponding points from the point cloud (frustum). The frustum is then passed through a PointNet \cite{qi2017pointnet} style architecture for classification of the frustum and box prediction.

PointPillars \cite{lang2019pointpillars} use a point cloud to pseudo-image transformation to be able to apply 2D convolutions for faster object detection. While the novel pillar encoding proposed in the paper helps reducing negative effects of point cloud's sparsity, it adds another layer of pre-processing. They use Single Shot Detector (SSD) \cite{liu2016ssd} style detection heads which use anchor boxes and non-maximum suppression, both of which are computation intensive. Fusion approaches like MV3D \cite{chen2017multi} and Liang et al. \cite{liang2018deep} make use of different views of the point cloud and camera images for feature fusion and object detection detection.

Recently, BirdNet+ \cite{barrera2020birdnet+} proposed object detection using BEV images of point clouds. This is the closest approach to our paper. Their network uses a two stage approach where they use a Faster R-CNN backbone for region proposals followed by box regression.  Pixor \cite{yang2018pixor} proposes a single stage detection model to predict oriented bounding boxes for objects using BEV images.



\subsection{Object detection using key points}

Zhou et al. \cite{zhou2019objects} demonstrate the object detection using key points approach by defining a key point as the image's center and regressing other parameters, such as offset to actual center, bounding boxes etc. SMOKE \cite{liu2020smoke} extends this concept to monocular 3D box detection in images on the KITTI dataset \cite{geiger2013vision} where a key point is defined as the object's center on the ground plane, in camera coordinates, projected to the image plane. They use a deep layer aggregation (DLA) \cite{yu2018deep} style network as backbone for feature extraction and three detection heads for key point detection and box parameter regression. 


\section{Proposed Method}

\subsection{BEV Representation}
3D point cloud is the standard representation of LiDAR data. This representation is sparse and irregular and thus inefficient for parallel processing on embedded systems. Range images and BEV images are equivalent 2D representations. In this work, we make use of BEV images as an efficient way of representing LiDAR data in a 2D grid. One key advantage is the distinct and non-overlapping location of objects in BEV space. We leverage this aspect for accurate localization of object centers. BEV space is commonly used for motion planning and thus it is better to localize in this space than in range images. 
For BEV image construction, we make use of the method used in BirdNet+ \cite{barrera2020birdnet+} with a resolution of $512\times256$. They use two input planes for encoding height and occupancy. 
Maximum height of all the 3D points mapped to the corresponding pixel in the BEV image is used. 
Occupancy plane's value is 0 if there was no 3D point mapped to it and 1 otherwise. This channel enables the network to explicitly mask invalid regions. We use an additional channel to encode intensity of LiDAR data.

\subsection{Network Architecture}

\begin{figure*}[t]
        \centering
        \includegraphics[width=\textwidth]{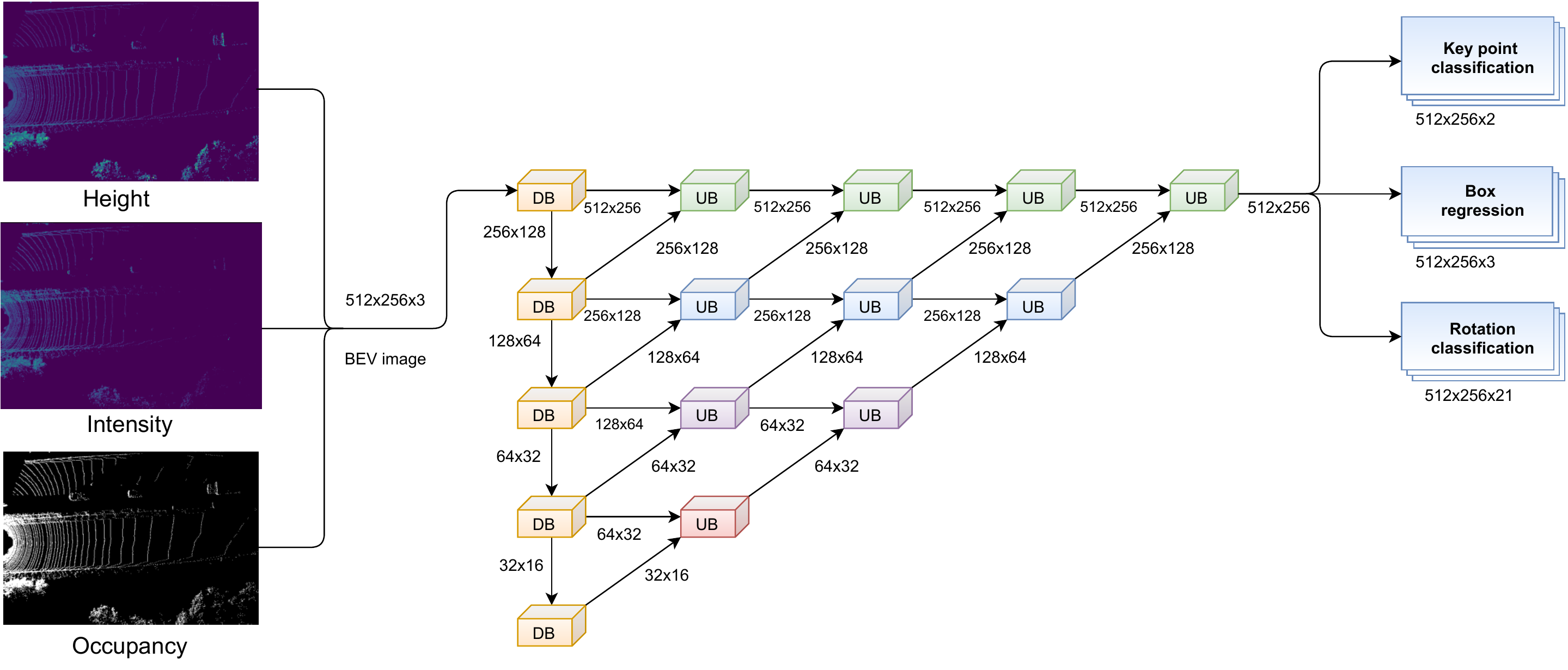}

        \vspace{3mm} 
        
         \includegraphics[trim={2.35cm 0 1.46cm 0},clip, width=\textwidth]{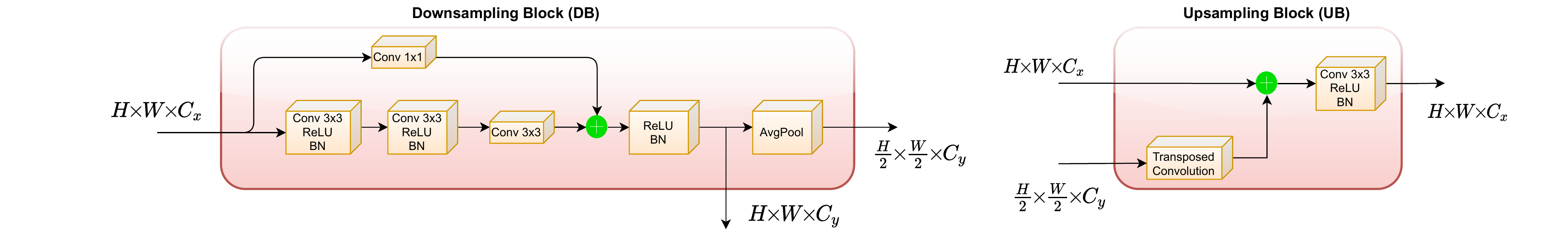}

        \caption{BEVDetNet Network architecture. DB and UB blocks in the top architecture are expanded below. {The number of filters in each DB level increases by a factor of 2 from top to bottom, starting at 32 channels for topmost DB and 512 for bottom most DB. All blocks at the same DB level have the same number of feature maps.}}
        \label{fig:Figure_overall_nw_bd}         
    
        
    \end{figure*}

The BEVDetNet architecture consists of the segmentation network for feature extraction and three prediction heads as shown in Figure~\ref{fig:Figure_overall_nw_bd}. 
{We adapt deep layer aggregation (DLA) \cite{yu2018deep} style network to provide feature maps at different spatial resolutions and at each level.
Internally, each feature extraction block is a modified residual unit with dilated convolutions augmented by large kernel context aggregation modules (CAM) \cite{wu2019squeezesegv2}. We call this the hybrid DLA architecture.}

Semantic segmentation architectures are computationally intensive \cite{briot2018analysis} and we carefully design an efficient version .  The prediction heads are key point classification (object center and class), box dimension regression for each key point group (length, width and height) and rotation prediction as a binned classification task for each group of key points.
The proposed model essentially modifies the semantic segmentation network to classify only pixels that are key points in an object. An additional benefit of this is that other segmentation classes like road and vegetation can be added without needing any additional computation. 
Here, we define the key point as the object center on the ground plane. KITTI dataset provides the object centers in camera coordinates. We first transform it to LiDAR coordinates and then to pixel coordinate of the BEV image.

    Key point classification can benefit from larger receptive field kernels which can provide better context aware feature extraction.
    This can be achieved efficiently using dilated convolutions \cite{hamaguchi1709effective} which increase receptive field of filters without increasing the number of parameters. Average pooling is better suited to preserve the overall context features compared to max pooling which picks the most dominant feature in the receptive field. Thus we propose to integrate these blocks into a traditional ResNet \cite{he2016deep}. 
    There are many unoccupied pixels in the BEV image due to limited resolution, reflections and other sources of noise. We observe that the unoccupied pixels severely degrade the representational ability of the initial convolution layers. We propose to make use of the Context Aggregation Module (CAM) defined in SqueezesegV2 \cite{wu2019squeezesegv2}. 
    {We modify it to use a larger kernel average pooling instead of max pooling to reduce the chances of encountering unoccupied windows}.

    
    The encoder consists of five feature extraction and downsampling blocks (DB) and the decoder consists of an equal number of feature aggregation and upsampling blocks (UB).
    As shown in Figure \ref{fig:Figure_overall_nw_bd}, each downsampling block uses a ResNet \cite{he2016deep} style unit that is modified to produce two outputs, one at the same spatial resolution as the input and the other at half resolution. Average pooling is used to downsample the input features.
    We make use of the CAM modules only in the first three downsample blocks as the receptive field of later blocks is larger and the effect unoccupied pixels is less pronounced. To further improve the receptive fields of the filters without increasing parameter size, we use dilated convolutions throughout the downsampling blocks. Hamaguchi et al. \cite{hamaguchi1709effective} showed that dilated convolutions are useful in feature extraction for small and crowded objects. This feature suits our problem of object detection in BEV images where objects are relatively small.

The upsampling blocks (UB) use transposed convolution followed by a normal convolution-relu-batch normalization stage to collect lower and higher level features from the corresponding hierarchies and fuses them by concatenation or summation. It is illustrated in Figure \ref{fig:Figure_overall_nw_bd}. The lower resolution feature maps are upsampled several times till it reaches the original BEV image resolution. 
Figure \ref{fig:Figure_overall_nw_bd} illustrates the exact architecture, number of layers and resolutions at each stage. The input and output dimensions are at the same resolution of $512\times256$.

\subsection{Loss Function}

The loss function is a weighted sum of losses corresponded to key point classification $L_{keypoints}$, box dimension regression for each key point $L_{box}$ and rotation classification task for each group of key points $L_{rotation}$ as shown below. The loss weights are tunable hyperparameters initialized to 1, 0.98 and 0.95. 
\begin{equation} \label{eq1}
    L_{total} = w_1 * L_{keypoints} + w_2 * L_{box} + w_3 * L_{rotation}
\end{equation}

\textbf{Keypoint classification}
Weighted crossentropy is used for keypoint classification as shown in Equation (2). Weights $w_c$ for the class $c$ is inversely proportional to its frequency $f_c$. This handles the large class imbalance problem in BEV images since most of the pixels do not correspond to objects of interest. Number of channels in the output feature map is the same as the number of classes. In our experiments for car detection, it is two channels. { Ground truth class for a pixel is $y_c$ and $\hat{y_c}$ is the predicted class. The loss weighted cross-entropy loss is then given as}
\begin{equation} \label{eq2}
    L_{keypoints} = -\sum_{c=1}^{C}w_cy_c\log{\Hat{y_c}}, \quad w_c = \frac{1}{\log(f_c + \epsilon)}
\end{equation}

\textbf{Box Regression} The box dimensions prediction output has 3 channels one each for length, width and height. Smooth L1 loss used is provided in Equation (3). The loss is computed only for pixels that have an object (positive samples). To normalize regression outputs to [0, 1], we construct the box regression target as \( \left[ \log(H),  \log(W), \log(L) \right] \). Since the input and output feature maps are of the same dimension, no offset adjustment is needed for centers and boxes. The predicted key points directly provide the object centers in the BEV image which can then be transformed directly to 3D centers. The resolution of the object center is limited to the BEV cell resolution and an additional center offset regression head can be used to refine. In our experiments, it did not provide significant improvements.

{
\begin{equation} \label{eq3}
    {L_{box}} =
    \begin{cases}
    0.5x^{2} & |x| < 1,\\
    |x| - 0.5 & otherwise
    \end{cases}
\end{equation}} where $(x = prediction - ground truth)$ for height, width and length predictions. \\

\textbf{Rotation as binned classification}
KITTI dataset provides rotation of boxes in the range \([-\pi, \pi]\) radians about the y-axis in camera coordinates.
Methods like Pixor \cite{yang2018pixor} predict rotation of bounding boxes as a regression of the sine and cosine components of the angle. 
Classification task suits our segmentation based architecture better and thus we
propose to predict rotation as a binned classification using a weighted cross-entropy loss like the key point classification head. We experimentally find that the binned classification  performs significantly better than using regression of sine and cosine of the angle.

To minimize the number of classes of rotation angles, we map all the box rotations to the positive range of [0, 180] degrees. Negative angles are converted to positive values by adding 180 to it as shown in Equation (4). For example, the orientation produced by +45 degrees is same as -135 degrees. This assumes heading direction is not needed.
We uniformly divide into 20 bins of 9 degrees each and have an additional bin for background. 
The rotation task is now classification of each group of key points and the corresponding receptive fields to one of these 21 bins. 
PointRCNN \cite{shi2019pointrcnn} performs a binned classification as a first step and refine it with an additional regression step. In our work, we directly do the final classification without needing any refinement and we exploit the rotational symmetry of box orientation to reduce the number of bins.

\begin{equation} 
\phi_{new}=\begin{cases}
			\phi, & \text{if $\phi$ \(>= 0\)}\\
            180 + \phi , & \text{otherwise}
		 \end{cases}
\end{equation} 

where \(\phi\) : original box rotation


\subsection{Post processing}
\begin{figure}[t]
    \centering
    \includegraphics[width=0.45\textwidth, height=0.45\textwidth]{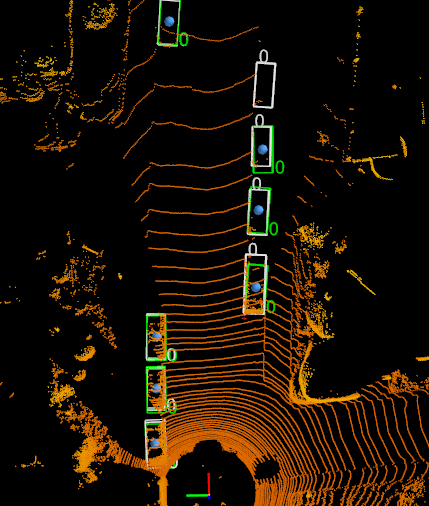}

    \caption{Illustration of object centers in 3D (denoted by {\color{blue}blue} dots) detected as key points and predictions of detected boxes (overlaid in {\color{green}green}), and ground truth is shown as white boxes.}
    \label{fig:Figure_keypoints_3d}
\end{figure}

The network predicts key points, boxes and rotation all in the BEV image. We make use of a $3x3$ window non-maximal suppression to obtain a unique key point from a group of neighboring key point predictions. It is performed using a simple distance based threshold and suppression of nearby key points assuming that the bounding boxes of two objects do not overlap. Finally, we convert the row and column indices $(r, c)$ of the predicted key point in the BEV image to 3D Cartesian coordinates $(x, y, z)$ in the point cloud using the equation below.
\begin{align}
    x = x_{max} - r * \delta , \quad
    y = y_{max} - c * \delta 
\end{align}

where \(\delta\) is the cell resolution of BEV which is set to $0.1m$. The z-coordinate can be obtained from the BEV image pixel value at $(r, c)$ as the pixel values contains the height. However, it is not needed for BEV precision score evaluation. Figure~\ref{fig:Figure_keypoints_3d} illustrates the final 3D key points and oriented bounding boxes.



\begin{table*}[htb]
\centering
\begin{adjustbox}{width=\textwidth}
\begin{tabular}{|l|c|c|c|c|c|c|c|c|c|c|} 
\hline
 & \multicolumn{2}{c|}{\textbf{Latency [ms/frame]} $\mathbf{\downarrow}$ } & \multicolumn{4}{c|}{\textbf{AP at IoU=0.5} $\mathbf{\uparrow}$ } & \multicolumn{4}{c|}{\textbf{ AP at IoU=0.7 } $\mathbf{\uparrow}$} \\ 
\cline{2-11}
 \textbf{Model} & \multicolumn{1}{c|}{\begin{tabular}[c]{@{}c@{}}Desktop \\ GPU$^1$ \end{tabular}} & \multicolumn{1}{c|}{\begin{tabular}[c]{@{}c@{}}Nvidia \\ Xavier$^2$ \end{tabular}} & \multicolumn{1}{c|}{Easy} & Moderate & \multicolumn{1}{c|}{Hard} & Average & Easy & Moderate & Hard & Average \\ 
\hline\hline
BirdNet+ \cite{barrera2020birdnet+} & 100 & -- & -- & -- & -- & -- & 84.80 & 63.33 & 61.23 & 65.14 \\ 
\hline
PointRCNN \cite{shi2019pointrcnn} & 100 & 747 & 90.20 & 89.63 & 89.39 & 89.58 & 92.13 & 87.39 & 82.72 & 85.66 \\ 
\hline
Part-A2 \cite{shi2019part} & 80 & 540 & 90.76 & 89.22 & 88.46 & 89.04 & 90.42 & 88.61 & \textbf{87.31}  & 88.19 \\ 
\hline
Pixor \cite{yang2018pixor} & 66 & 372 & 89.62 & 83.45 & 80.11 & 82.59 & 86.79 & 80.75 & 76.60 & 79.46 \\ 
\hline
Second \cite{yan2018second} & 50 & 165 & \textbf{90.79} & 90.14 & \textbf{89.51} & \textbf{89.90} & 89.96 & 87.07 & 79.66 & 83.71 \\ 
\hline
SA-SSD \cite{he2020structure} & 40 & -- & -- & -- & -- & -- & \textbf{95.03}  & \textbf{91.03}  & 85.96 & \textbf{89.00} \\ 
\hline
PointPillars \cite{lang2019pointpillars} & 16 & 101 & 90.78 & \textbf{90.18} & 89.46 & 89.89 & 88.32 & 86.10 & 79.83 & 83.23 \\ 
\hline\hline
\textbf{BEVDetNet (ours)}  & \textbf{3}  & \textbf{6}  & 87.82 & 87.78 & 87.25 & 87.51 & 82.46 & 77.90 & 77.45 & 78.28 \\
\hline
\end{tabular}
\end{adjustbox}
\caption{Average precision (AP) and computation latency for car detection measured on KITTI's validation set at two different Intersection over Union (IoU) settings. $^1$We used a laptop GPU while others use desktop GPUs of similar performance. $^2$It was not possible to convert other models to TensorRT due to dynamic input sizes, usage of unsupported CUDA kernels and complexity of architectures. All models are in FP32 resolution to enable a fair comparison. }
\label{tab:Table_AP_latency}
\end{table*}

\section{Experimental Evaluation}

\subsection{Experimental Setup}

We generate BEV images from KITTI object detection dataset's 7,481 point cloud samples following BirdNet+ \cite{barrera2020birdnet+}. We divide these into training and validation sets following the split used in PointPillars \cite{lang2019pointpillars}. We limit the range (meters) of the points to $x = (0, 60)$, $y = (-30, 30)$ and $z = (-2.73, 1.27)$. The height and width of the BEV for training are cropped to 512 and 256 respectively. The pixel values are the z-coordinates clipped to the range of $(-2.73, 1.27)$ and offset by 2.73 to bring the range of pixel values to $(0, 4)$ and further normalized to the range (0, 1).
We use extensive data augmentation, i.e., horizontal flips, global rotation of 3D point cloud in the range (-5, 5) degrees and sampling based object augmentation  \cite{yan2018second}, increasing the training set to 16,000 samples.

The network was trained end-to-end using the Adam optimizer. We evaluated batch sizes of $(8, 16, 24, 32, 48)$ and found a batch size of 16 to yield the best performance. 
Networks were trained for $(50, 60, 80, 120)$ epochs and 50 epochs yields the best performance.
Unlike other methods, we do not use any pre-trained weights and the network is trained from scratch.
The training takes approximately 30 hours on two Titan RTX GPUs with a combined memory of 48 GBs.

\subsection{Metrics and performance comparison}

Table \ref{tab:Table_AP_latency} shows quantitative results on the KITTI evaluation dataset. For practical purposes, IoU of 0.5 is sufficient for object detection. Thus we focused on tuning our model for this metric. We additionally report accuracy for IoU of
0.7. As the focus of our work is on real-time models, we also provide a comparison of computational latency on both a {laptop} GPU and on an embedded Nvidia Jetson Xavier AGX platform using the OpenPCDet \cite{openpcdet2020} library. With its 8-core ARM processor and 512-core GPU, the Xavier AGX can provide up to 32 Tera Operations per Second (TOPS) compute performance at a small form factor and hence can serve as a representative platform for autonomous driving applications. It was not possible to benchmark SA-SSD \cite{he2020structure} and BirdNet+ \cite{barrera2020birdnet+} on the Xavier development kit. SA-SSD dependency library compilation failed and BirdNet+ uses ROS \cite{ros} packages for BEV generation which required additional effort to port. TensorRT is an optimizing compiler for Nvidia Xavier and it was not possible to convert other models to TensorRT due to dynamic input sizes, usage of unsupported CUDA kernels and complexity of architectures. 

BEVDetNet outperforms all the other methods in terms of inference latency time by a large margin of 17X on the embedded GPU and 3X on the laptop GPU. Inference was executed on a laptop with Nvidia RTX2080 MaxQ GPU and Nvidia TensorRT engine. Most other models like PointPillars were benchmarked on a Nvidia 1080Ti desktop GPU. Both GPUs have comparable TFLOPS. We observe that other methods have a large variations in their latency due to scene dependent variations. 
Our accuracy for IoU of 0.5 is promising as it is only 2\% lesser than the best model. The gap is larger for IoU of 0.7 and the main reason might be due to larger complexity needed for higher precision localization. 
The closest to our approach is the recent BirdNet+ \cite{barrera2020birdnet+} and we significantly outperform by 12\% for IoU of 0.7.


\subsection{Ablation studies}



\begin{figure*}[h]
    \begin{subfigure}{\textwidth}
    \centering
    \includegraphics[width=\textwidth, trim={0  0 0 1.3cm },clip]{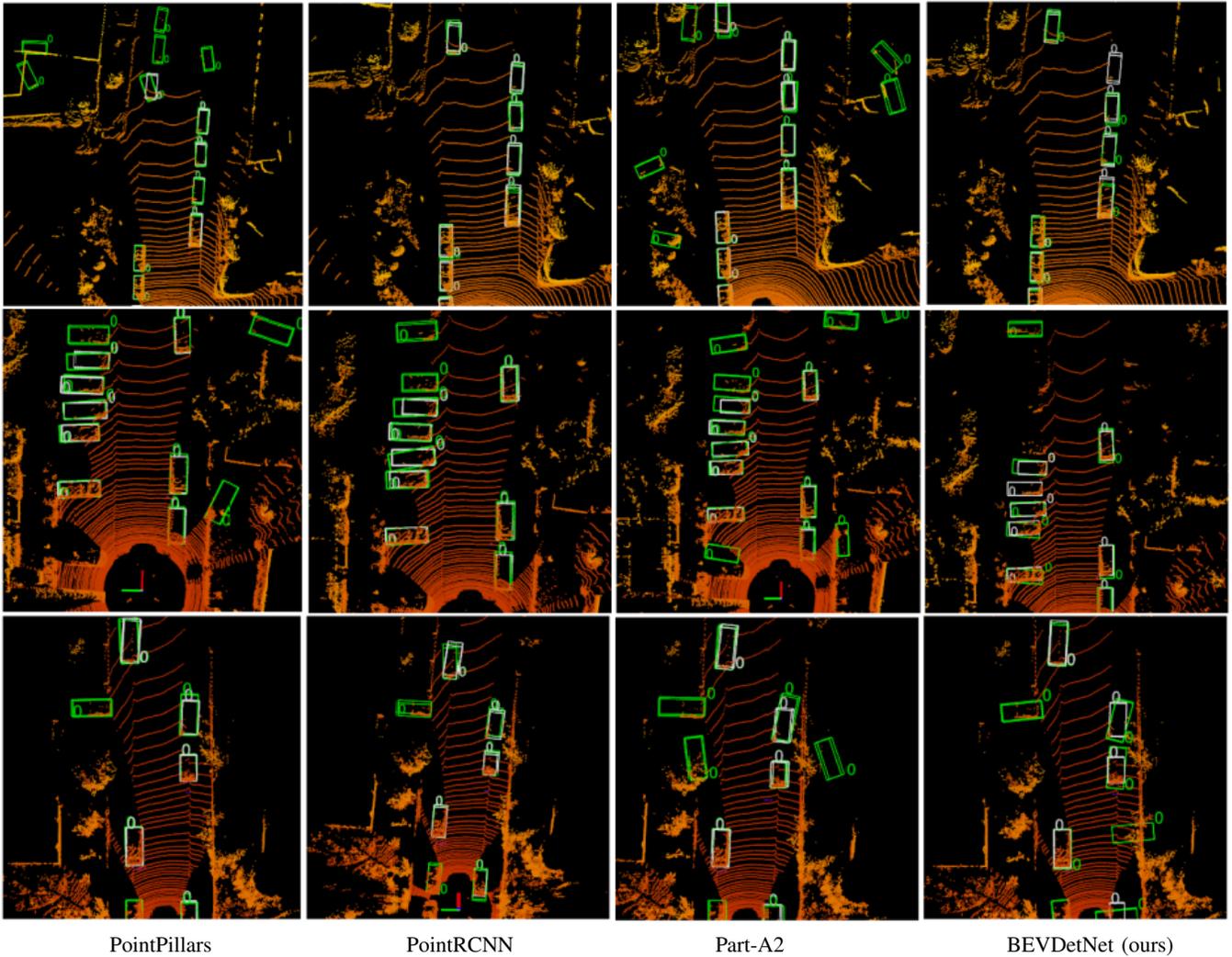}  
    \subcaption*{\hspace{0.7cm} PointPillars \hspace{2.6cm}   PointRCNN \hspace{2.7cm} Part-A2 \hspace{3cm} BEVDetNet (ours)}
    \end{subfigure}
    \caption{Qualitative comparison of results. {\color{green}Green} and white boxes illustrate prediction and ground truth respectively. }
    \label{fig:Figure_qualitative}
\end{figure*}


\begin{table}[t]
\centering

\begin{adjustbox}{width=0.49\textwidth}
\begin{tabularx}{\columnwidth}{|X|c|c|c|c|}
\hline
 & \textbf{Latency} & \multicolumn{3}{c|}{\textbf{AP}} \\ \cline{3-5} 
\textbf{Backbone} & [ms/frame] & {Easy} & {Moderate} & {Hard} \\ \hline\hline
PSPNet & 32 & 72.80 & 68.60 & 66.40 \\ \hline
U-Net baseline & 5 & 69.74 & 69.42 & 66.28 \\ \hline
Hybrid-DLA (ours) & \textbf{3} & \textbf{82.46} & \textbf{77.90} & \textbf{77.45} \\ \hline
\end{tabularx}
\end{adjustbox}
\caption{Ablation study with different encoder-decoder architectures as backbone at IoU=0.7. }
\label{tab:Table_ablattion_backbone}


\vspace{3mm}

\begin{adjustbox}{width=0.49\textwidth}
\begin{tabular}{|l|l|l|l|l|}
\hline
\multicolumn{1}{|c|}{\multirow{2}{*}{\textbf{Classification Loss}}} & \multirow{2}{*}{\textbf{Regression Loss}} & \multicolumn{3}{c|}{\textbf{AP}} \\ \cline{3-5} 
\multicolumn{1}{|c|}{} &  & \textbf{Easy} & \textbf{Moderate} & \textbf{Hard} \\ \hline\hline
Focal & L1 & 78.27 & 75.04 & 72.76 \\ \hline
Cross-entropy & L1 & 75.17 & 74.17 & 72.75 \\ \hline
Focal & Smooth L1 & 80.02 & 76.23 & 75.12 \\ \hline
Cross-entropy & Smooth L1 & \textbf{82.46} & \textbf{77.90} & \textbf{77.45} \\ \hline
\end{tabular}
\end{adjustbox}
\caption{Performance comparison with different losses at IoU=0.7. }
\label{tab:Table_ablation_loss}


\vspace{3mm}


\begin{adjustbox}{width=0.4\textwidth}
\centering
\begin{tabular}{|c|c|c|c|}
\hline
\textbf{Quantization} & \textbf{\begin{tabular}[c]{@{}c@{}}Latency\\ (ms)\end{tabular}} & \textbf{\begin{tabular}[c]{@{}c@{}}Model size\\ (MB)\end{tabular}} & \textbf{\begin{tabular}[c]{@{}c@{}}AP\\ (average)\end{tabular}} \\ \hline
FP32 & 6 & 59 & \textbf{87.51} \\ \hline
FP16 & 4 & 39 & 86.44 \\ \hline
INT8 & \textbf{2.1} & \textbf{22} & 84.20 \\ \hline
\end{tabular}
\end{adjustbox}
\caption{Performance comparison of different model resolutions at IoU = 0.5}
\vspace{-2mm}
\label{tab:ablation-quantization}
\end{table}


Table~\ref{tab:Table_ablattion_backbone} shows an ablation study of different segmentation architectures. Firstly, we make use of the state of the art semantic segmentation network for images namely PSPNet \cite{zhao2017pspnet}. We also make use of a simple U-Net baseline which has the same topology as our proposed network with simple convolution layers. We demonstrate that the proposed Hybrid-DLA architecture provides a large improvement in performance with improved latency. We also show a large improvement over PSPNet using design techniques which handle sparsity of BEV images. 

Furthermore, we evaluate weighted cross-entropy loss and focal loss \cite{lin2017focal} (cp. Table~\ref{tab:Table_ablation_loss}) for the key point classification task and L1 and Smooth L1 loss for the box regression task. Smooth L1 provides a significant improvement over simple L1 loss.
A combination of weighted cross-entropy and Smooth L1 provided the best performance.

{
We also experiment with reduced precision quantized models for 16-bit floating point (FP16) and 8-bit fixed point integer (INT8) as shown in Table \ref{tab:ablation-quantization}. This is along the lines of our goal to improve efficiency for embedded deployment. FP16 precision model was created using the quantization option in TensorRT while the INT8 model was converted using a calibration dataset made from the training BEV images.
The latency improved significantly reaching \textit{2 ms} for INT8 resolution. There is a noticeable degradation in accuracy as we are doing offline quantization but this gap can be reduced using quantization aware training.
 }

\subsection{Failure Cases}
Typical qualitative results of our method comparing other methods are provided in Figure~\ref{fig:Figure_qualitative}. We observe two main failure modes that lead to lesser AP score.

\textbf{Missed detections} occur mainly when there object points are sparse with few pixels available in the BEV image. While we could train with more such samples using data augmentation, it was experimentally found that this leads to an increase in false predictions, which we wanted to avoid. Handling such cases using pre-processing steps like depth completion is a promising direction.
    
\textbf{Incorrect object center predictions} occur due to the reconstruction process from 2D key points to 3D centers. Due to the simplicity of our method in selecting a single center pixel from the group of pixels for a key point, sometimes a incorrect neighboring pixel is selected as center. Since each pixel represents $0.1m$ on the ground, even an offset by 2 pixels results in a significant offset in the 3D box center. One straightforward solution is to improve the cell resolution in the BEV image resulting in lesser errors in case of offsets in center prediction. But it would result in a higher resolution BEV image and significantly increase the latency.

\section{Conclusion}

In this paper, we developed a low latency simple architecture for 3D object detection using BEV images of LiDAR point clouds. We showed competitive performance on KITTI dataset and achieved a latency of 4 ms on an automotive embedded platform namely Nvidia Xavier. Our architecture is based on semantic segmentation unifying key point detection for object centers, box predictions and rotation classification. The architecture can directly benefit from recent improvements of semantic segmentation algorithms developed for images. In future work, we plan to design a scalable family of models which can support different accuracy-latency tradeoff values. 

\section*{Acknowledgment}
\begin{wrapfigure}{r}{0.04\textwidth}
\includegraphics[width=7mm]{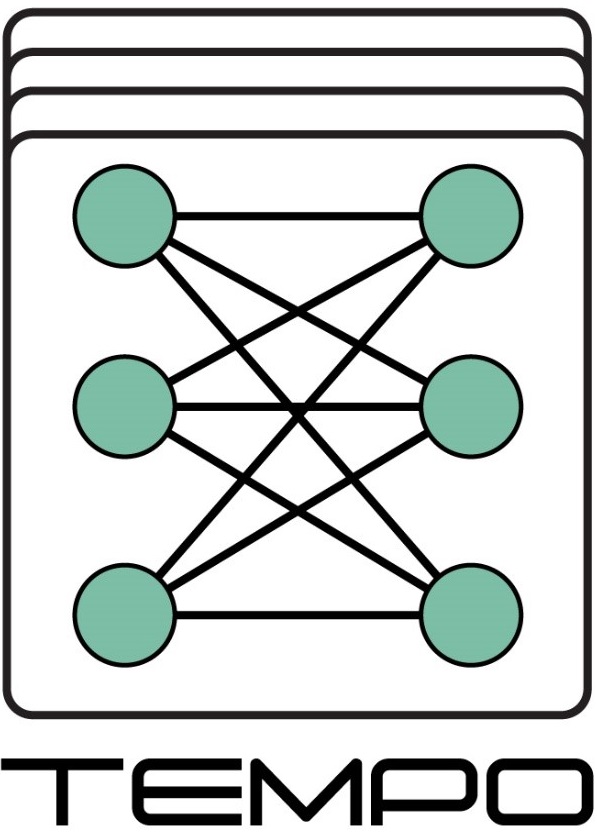}
\end{wrapfigure}
We are funded by the Electronic Components and Systems for European Leadership Joint Undertaking grant No 826655 receiving support from the European Union’s Horizon 2020 research and innovation programme. Further partial funding is provided by the German Federal Ministry of Education and Research.


{\small
\bibliographystyle{IEEEtran}
\bibliography{egbib}
}

\end{document}